%% file: main.tex
\def\year{2022}\relax
\documentclass[letterpaper]{article} 
\usepackage{aaai22}  
\usepackage{times}  
\usepackage{helvet}  
\usepackage{courier}  
\usepackage[hyphens]{url}  
\usepackage{graphicx} 
\usepackage{natbib}  
\usepackage{caption} 
\DeclareCaptionStyle{ruled}{labelfont=normalfont,labelsep=colon,strut=off} 
\usepackage{algorithm}

\usepackage{newfloat}
\usepackage{listings}

\usepackage{epsfig}
\usepackage{amsmath}
\usepackage{amssymb}
\usepackage{adjustbox}
\usepackage{wrapfig}
\usepackage{multirow}
\usepackage{algpseudocode}
\usepackage{arydshln}

\usepackage{subfigure}

\usepackage{color}
\usepackage{rotating}

\floatstyle{ruled}
\newfloat{listing}{tb}{lst}{}
\floatname{listing}{Listing}
\newcommand{\etal}{et~al.}

\title{FrePGAN: Robust Deepfake Detection Using Frequency-level Perturbations}
\author{Yonghyun Jeong\textsuperscript{\rm 1}, Doyeon Kim\textsuperscript{\rm 1}, Youngmin Ro\textsuperscript{\rm 1}, Jongwon Choi\textsuperscript{\rm 2*}\\
}

\vspace{-0.90mm}
\affiliations {
    \textsuperscript{\rm 1}Samsung SDS, Seoul, Korea\\
    \textsuperscript{\rm 2}Department of Advanced Imaging, Chung-Ang University, Seoul, Korea\\
    \small
    \{yonghyunjjj, ddkim.0131, youngmin4920\}@gmail.com ,\ choijw@cau.ac.kr
} 

\begin{document}

\maketitle

\input{0-abstract}

\input{1-introduction}
\input{2-related_work}
\input{3-method}
\input{4-result}
\input{5-conclusion}

\section{Acknowledgment}
This work was supported by Samsung SDS and Institute of Information \& communications Technology Planning \& Evaluation (IITP) grant funded by the Korea government(MSIT) (2021-0-01341, Artificial Intelligence Graduate School Program(Chung-Ang University); 2021-0-01778, Development of Human Image Synthesis and Discrimination Technology Below the Perceptual Threshold; 2021-0-02067, Next Generation AI for Multi-purpose Video Search).
{
\bibliography{aaai22}
}
\end{document}

%% file: 0-abstract.tex
\begin{abstract}
Various deepfake detectors have been proposed, but challenges still exist to detect images of unknown categories or GAN models outside of the training settings.  Such issues arise from the overfitting issue, which we discover from our own analysis and the previous studies to originate from the frequency-level artifacts in generated images. We find that ignoring the frequency-level artifacts can improve the detector's generalization across various GAN models, but it can reduce the model's performance for the trained GAN models. Thus, we design a framework to generalize the deepfake detector for both the known and unseen GAN models. Our framework generates the frequency-level perturbation maps to make the generated images indistinguishable from the real images. By updating the deepfake detector along with the training of the perturbation generator, our model is trained to detect the frequency-level artifacts at the initial iterations and consider the image-level irregularities at the last iterations. For experiments, we design new test scenarios varying from the training settings in GAN models, color manipulations, and object categories. Numerous experiments validate the state-of-the-art performance of our deepfake detector.
\end{abstract}

%% file: 1-introduction.tex
\begin{figure}[t] 
\includegraphics[width=1\linewidth]{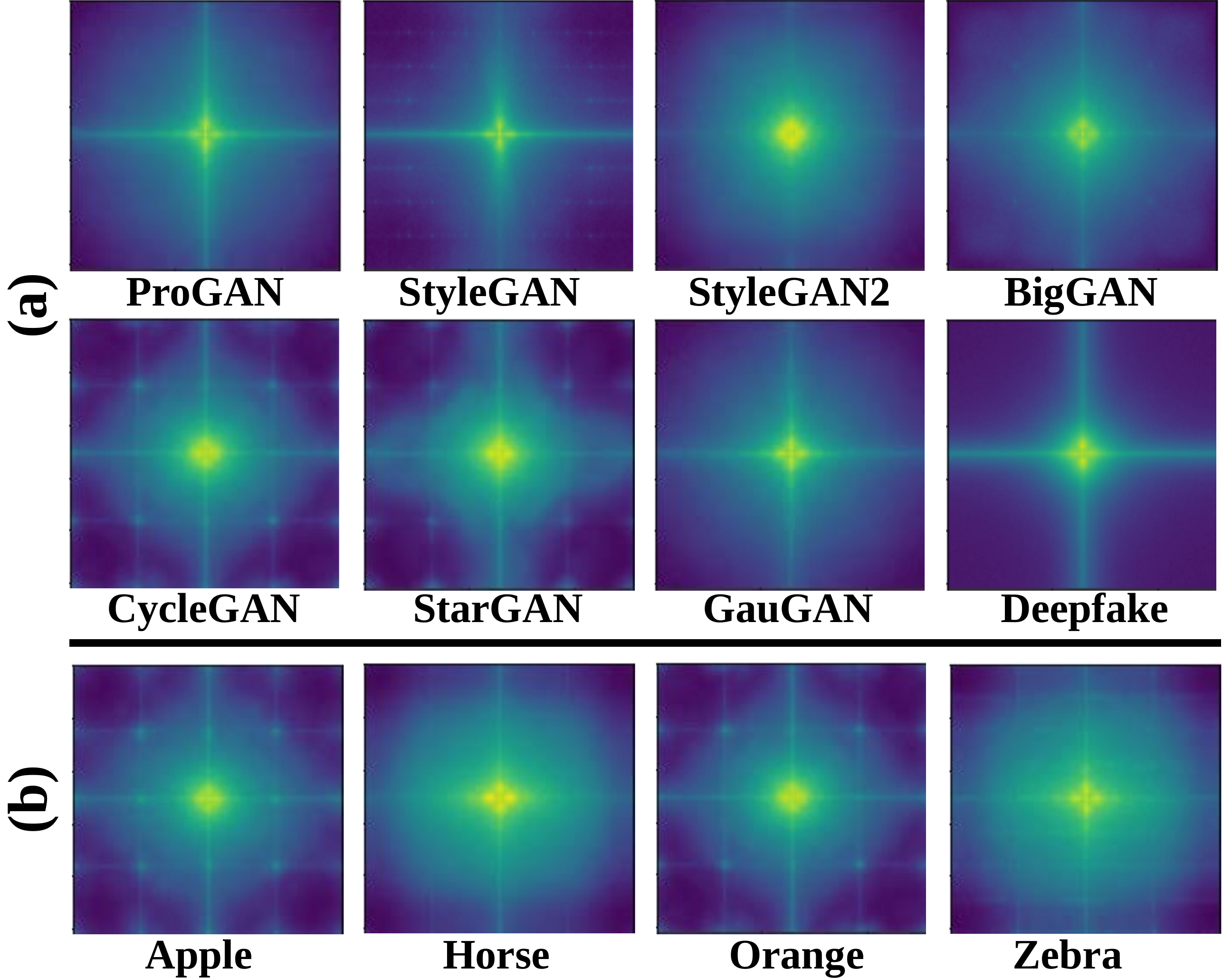}
\caption{\textbf{Various Patterns of Frequency Artifacts.} The frequency-level artifacts can be extracted by averaging the frequency-level spectrum of the generated images. The appearance of artifacts is evident but uniquely vary by the type of GAN model or object category. Thus, we can analyze that the artifacts are easily detected by their evident appearances but can cause overfitting to the training settings due to their uniqueness. 
Thus, the artifacts should be ignored for generalized detection, but they are still useful for detection of specific GAN models.} 
\label{fig:2d_spec} 
\end{figure}
\section{Introduction} 
The recent rise of Generative Adversarial Networks (GAN)~\cite{gan,progan,stylegan,stylegan2} has allowed the easy and extensive generation of highly realistic fake images, as known as deepfakes. 
Unfortunately, the risk of malicious abuse of deepfakes also rises with such an advancement~\cite{nguyen2019deep}, and the importance of detecting deepfakes has become crucial. 
The target range of deepfakes has broadened from swapping the face of the celebrity on the body of pornography to spreading misinformation in social media as fake news, and even alluring victims to transfer money for scams~\cite{deepfakesandbeyond, nguyen2019deep}.
To solve this issue, the tech giants and the academia have joined together for `Deepfake Detection Challenge'~\cite{DFDC, DFDC2020} to promote the current issues and encourage fellow researchers to tackle this problem. 

As confirmed by several previous studies~\cite{chen2020ssd}, the CNN-based generative models are known to have limitations in reconstructing the high-frequency components. 
However, as shown in Fig.~\ref{fig:2d_spec}, although the frequency-level artifacts are effective to detect the generated images for the specific GAN models, it is easy for the detectors to be overfitted to the training settings, due to the unique appearances of the frequency-level artifacts varying by the CNN structures and training categories.
Thus, it can be analyzed that the frequency-level artifacts are effective to detect the generated images from the known GAN models, but the key to the generalization of deepfake detectors is to reduce the effect of frequency-level artifacts during training. 

{\let\thefootnote\relax\footnotetext{$^{*}$Corresponding author}}
Based on intuition, we propose a novel framework composed of two modules: the Frequency Perturbation GAN (FrePGAN) and the deepfake classifier. 
FrePGAN contains the frequency-level perturbation generator and the perturbation discriminator, which cooperate to adversarially generate perturbation maps added onto both the real and fake images for reduced differences in the frequency-level.
Then, the perturbed images are fed into the deepfake classifier to distinguish the fake images from the real ones in the pixel-level. 
To train the deepfake detector to utilize both the frequency-level artifacts and the general pixel-level irregularity, we update the frequency-level perturbation generator and the deepfake classifiers alternatively.
To validate the performance of our model, we conduct numerous experiments using multiple deepfake datasets. 
Including the benchmark evaluations, we have designed three types of distinct test settings using unknown categories, models, and manipulations unused during training.
Our model achieves state-of-the-art performance in both the known and unseen settings. 

Our paper makes the following contributions:
\begin{itemize}
\item We develop FrePGAN to generate the frequency-level perturbation maps to ignore the domain-specific artifacts in fake images.
\item The perturbation maps obtained from FrePGAN are added to the given input images, which can reduce the effect of domain-specific artifacts and improve the generalization ability of the deepfake detector.
\item FrePGAN and the deepfake classifier are updated alternatively to train the deepfake classifier to consider both the frequency-level artifact and the general feature.
\item Our model achieves superior results compared to the state-of-the-art models and robust detection performance of generated images in the known and unseen domains.
\end{itemize}

%% file: 2-related_work.tex
\section{Related Work}
The previous work can be categorized into the physiological feature-based, image-based, and frequency-based detection.

\subsection{Physiological Feature-based Detection}
With the rise of realistic human deepfakes, most studies focus on the temporal properties, such as facial features~\cite{agarwal_protecting_2019, matern, resnet, Montserrat}, incoherent head poses~\cite{headpose}, and lack of eye-blinking~\cite{eyeblinking}. \cite{faceforensics++, DFDC, Celeb_DF_cvpr20} provide large-scale datasets and evaluate various image forensics for face manipulations. 
However, since most of these methods focus on the face only, they can be ineffective in non-facial domains. 

\subsection{Image-based Detection }   
To expand the detection range, some studies take images as input data. Tralic~\etal~\cite{tralic} analyze the inconsistencies in blocking artifacts generated during JPEG compression~\cite{tralic}. 
Ferrara \etal~\shortcite{ferrara} explore the demosaicing artifacts generated in manipulated images due to a color filter array~\cite{ferrara} but the artifacts can disappear during resizing.
Thus, some focus on the deviations in lighting conditions to detect manipulations~\cite{carvalho, peng2}. 
Also, \cite{bayar} suggest learning the prediction error filters for generalization but it struggles with post-processing methods used to manipulated regions.
Thus, Cozzolino~\etal~\cite{cozzolino} propose an adaptable neural network to new target domains using a few training samples~\cite{cozzolino}. 
Wang~\etal~\cite{adobe} use RGB images to distinguish cross-model manipulations, such as blurring and JPEG~\cite{adobe}.
Also, \cite{guarnera} explore the hidden traces by analyzing the last computational layer to predict real and fake and the most probable technique used.
Recently, Zhao~\etal~\cite{zhao2021}suggests a multi-attention network to attend different local parts for the artifacts and aggregate the high and low features for classification~\cite{zhao2021}.

\subsection{Frequency-based Detection }
Some analyze the spectral traces in the frequency domain, as
\cite{kirchner} suggests using the frequency artifacts with the variance of prediction residue. 
Also, \cite{huang} employ Fast Fourier Transform~\cite{fft} and singular value decomposition to identify copy-move manipulations.
\cite{marra} suggest a GAN-specific detection using the artificial fingerprints in the frequency domain, and \cite{bappy} propose a manipulation localization architecture using spatial maps and frequency domain correlation. 
Also, \cite{frank} analyze the frequency artifacts using Discrete Cosine Transform, while \cite{zhang} exploit the artifacts induced by the up-sampler of GANs. 
Others~\cite{watch_cvpr20, unmasking} exploit the spectral distortions via azimuthal integration, while \cite{wacv} adopt the bilateral high-pass filters for generalized detection. 
Recently, \cite{ijcai} propose to re-synthesize testing images and extract visual cues for flexible detection.

%% file: 3-method.tex
\begin{figure*}[t]
\centering
{\includegraphics[width=15.5cm]{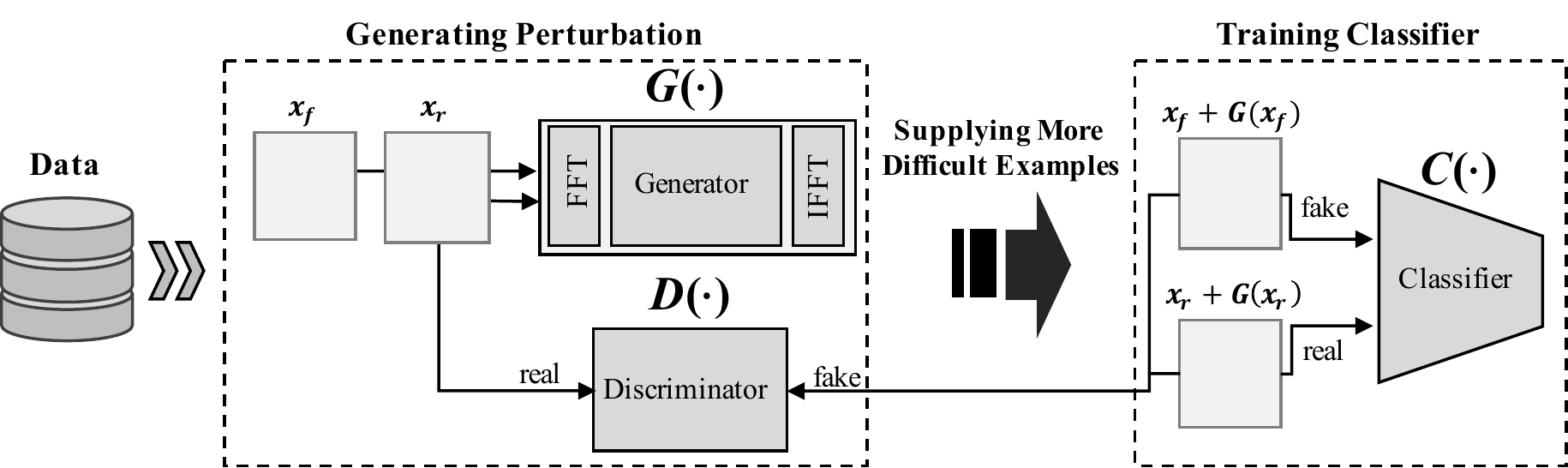}}
\caption{\textbf{Overall framework. }
Consisted of FrePGAN for generating the frequency-level perturbations and the deepfake classifier for distinguishing the real and fake, the framework allows detecting fake images with the balanced effect of domain-specific frequency-level artifacts and general image-level irregularity in the generated images.
}
\label{fig:framework}
\end{figure*}

\section{Deepfake Detection Framework}
We design a generalized deepfake detector containing the Frequency Perturbation GAN (FrePGAN) and the deepfake classifier.
FrePGAN generates the frequency-level perturbation maps for the deepfake detector to ignore the frequency-level artifacts.
To reduce the effect of the frequency-level artifacts, both real and fake images are added with the generated perturbation maps of FrePGAN, respectively. 
The deepfake classifier is designed to distinguish between the real and fake images. 
The visual illustration for the overall architecture is shown in Fig.~\ref{fig:framework}.

\subsection{Training of Deepfake Detection Framework}
Though FrePGAN and the deepfake classifier can be trained in a sequence, we purposefully train both networks in one iteration for comprehensive training of various properties of the perturbation maps.
Also, through the alternating update, we can enhance the generalization of the deepfake classifier by expanding the variety of its input data.
At the initial updates, FrePGAN fails to generate the proper perturbation map to ignore the effect of frequency-level artifacts, so the deepfake classifier is trained to distinguish the fake images from the real images by using the artifact that is easy to be detected.
On contrary, when FrePGAN is sufficiently trained to generate the perturbation maps confusing the real and fake images, the deepfake classifier needs to extract the new feature that works generally across the various types of GAN models.
As a result, the alternating updates can make the deepfake classifier consider the frequency-level artifacts and the general features simultaneously.

\subsection{Frequency Perturbation GAN}
To train FrePGAN, we build a novel architecture composed of two major parts: the perturbation map generator trained by the perturbation generation losses, and the perturbation discriminator.
The input of FrePGAN is an image $x\in\mathbb{R}^{w\times h\times c}$ where $w$, $h$, and $c$ present its width, height, and number of channels, respectively.
Each and every input image is labeled by $y$ either as \textit{real} ($y=0$) or \textit{generated} ($y=1$), representing either actually captured in the real world or generated by GAN.
We define the real and generated images as $x_r\equiv x_{y=0}$ and $x_f\equiv x_{y=1}$ respectively, and thus $x$ would be one of $x_r$ or $x_f$.
The perturbation map generator and the perturbation discriminator are denoted as $G(\bullet)$ and $D(\bullet)$, respectively. 

\begin{figure}[h!]
\centering
\includegraphics[width=0.95\linewidth]{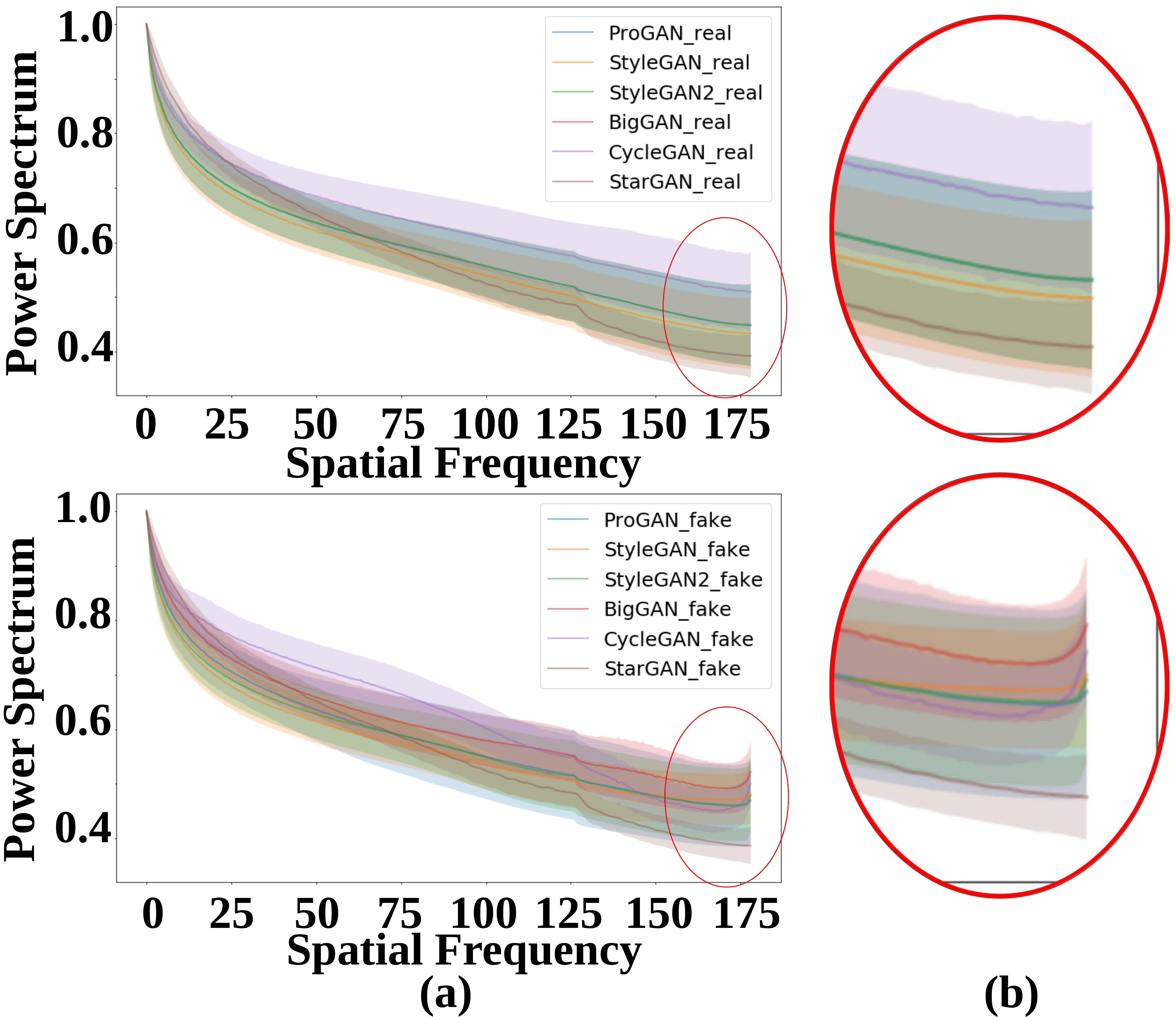}
\caption{\textbf{Comparison in power spectra of real and fake data} We average the power spectrum from (a) the entire training data and (b) the generated data, respectively. For every GAN model of the graph, the fake images suffer from the dramatic increment of the high-frequency components.}
\label{fig:1d_spec} 
\end{figure}

\subsubsection{Perturbation Map Generator}
As in Fig.~\ref{fig:1d_spec}, the real and fake images can be easily distinguished when transformed into the frequency domain. Also, it can be observed that the frequency-level artifacts mainly locate at the high-frequency components. Thus, by adding the frequency-level perturbations, we can reduce the effect of domain-specific artifacts.

To ignore the frequency-level artifacts, the perturbation should be generated in the frequency domain as well. Thus, we utilize the frequency map transformed from the original image as the input of the perturbation map generator.
The perturbation map generator contains three modules: frequency-level transformer, frequency-level generator, and the inverse frequency-level transformer.

First, the frequency-level transformer converts the input image into the frequency map by employing Fast Fourier Transform (FFT)~\cite{fft}, as denoted by $\widetilde{x}=\mathcal{F}(x)$ where $\widetilde{x}\in\mathbb{R}^{w\times h\times 2c}$ is the frequency map transformed from $x$ and $\mathcal{F}(\bullet)$ represents the operation of FFT.
The number of channels of $\widetilde{x}$ becomes doubled because each image channel is separated into two channels for the real and imaginary parts of the frequency-map. 

Then, the frequency-level generator receives $\widetilde{x}$ to generate a frequency map with the same size of $\widetilde{x}$.
The scheme of the generator is similar to those of image-to-image translation GANs~\cite{pix2pix_gan, cyclegan, disco_gan, stargan, stargan2, few_shot_img2img}, which contain the encoder and decoder.
Thus, when the frequency-level generator is denoted by $H$, and $\mathbf{\widetilde{z}}=H(\widetilde{x})$ where $\mathbf{\widetilde{z}}\in\mathbb{R}^{w\times h\times 2c}$ is the output from the frequency-level generator.

Lastly, the generated map $\mathbf{\widetilde{z}}$ is transformed into a pixel-level perturbation map. The overall operation of the perturbation map generator is derived with a given input of $x$ as:
\begin{align}
    G(x) = \mathcal{F}^{-1}( H( \mathcal{F}(x))),
\end{align}
where $\mathcal{F}^{-1}(\bullet)$ means the inverse FFT operation.
We need to remark that the final output from the perturbation map generator is shaped by $G(x)\in\mathbb{R}^{w\times h\times c}$, which is shaped by the same size of input $x\in\mathbb{R}^{w\times h\times c}$.

\subsubsection{Perturbation Discriminator}
To enhance the effect of the generated perturbation maps, we add the perturbation discriminator to adversarially train the perturbation map generator.
The overall architecture follows the conventional GAN discriminator~\cite{dcgan} that down-samples the input features by the consecutive convolution layers and performs binary classification at the last convolution layer.
By the last fully connected layer, the perturbation discriminator distinguishes the output of the perturbation map generator from the original image. 
Thus, for an input of $x$, the target prediction of the perturbation discriminator is a probability that can be represented by $D(x_{r}) = 0$ and $D(G(x)) = 1$.

\subsubsection{Training of FrePGAN}
The two compositions of FrePGAN are adversarially trained to generate the perturbation maps from the input images.
Thus, when real images are given to FrePGAN, the empty perturbation maps should be ideally acquired after the generator, due to the absence of frequency-level artifacts.
In contrast, when the perturbation maps are added to the fake images, the distribution of the added images should be difficult to distinguish from that of real images.

At every iteration, two training steps alternate, updating the perturbation map generator and the perturbation discriminator, respectively.
The perturbation map generator is updated by minimizing the perturbation generation loss ($\mathcal{L}_G$) while trying to maximize the discriminator loss ($\mathcal{L}_D$) for the update of the perturbation discriminator. 
Thus, the overall training of FrePGAN can be defined as:
\begin{equation}\label{eq:updateGAN}
    \begin{aligned}
        \widehat{G}, \widehat{D} = \arg_{G, D} \min_G \max_D \mathcal{L}_{G} + \mathcal{L}_{D}.
    \end{aligned}
\end{equation}
\normalsize

$\mathcal{L}_G$ has the generative adversarial loss ($\mathcal{L}_{adv}$) and the compression loss ($\mathcal{L}_{com}$) to compress the magnitude of perturbation maps.
At every mini-batch update, $G$ is first updated to minimize the following loss:
\begin{equation}\label{eq:updateG}
        \mathcal{L}_G = \lambda\mathcal{L}_{adv} + (1 - \lambda)\mathcal{L}_{com},
\end{equation}
where $\lambda$ is a hyperparameter to tune the scales of $\mathcal{L}_{adv}$ and $\mathcal{L}_{com}$. In this work, we use $\lambda = 0.5$.

We employ $\mathcal{L}_{adv}$ for the perturbation map generator to generate the perturbation maps added to the images to be indistinguishable from the real images by the perturbation discriminator.
Since $x_f$ is also the sample generated from other GAN models, $x_f$ is improper to be considered as the real sample for the adversarial training of FrePGAN.
Thus, 
\small
\begin{equation}\label{eq:advLoss}
\begin{aligned}
    \mathcal{L}_{adv} = \mathbb{E}_{x\sim \mathbf{X}}[\log\left(1 - D\left(G\left(x\right)\right)\right)],
\end{aligned}
\end{equation}
\normalsize
where $\mathbf{X}$ represents the batch sets of images.

If only the generative adversarial loss is considered during training, FrePGAN would not be able to preserve even the distribution of real images by adding a large magnitude of perturbation maps.
Since the purpose of employing FrePGAN is to obtain a similar distribution of real images, we additionally include the compression loss $\mathcal{L}_{com}$ in the perturbation generation loss.
Thus, 
\small
\begin{equation}\label{eq:comLoss}
    \mathcal{L}_{com} =\mathbb{E}_{x\sim \mathbf{X}}\left[\left \|G(x)\right\|_2^2\right]
\end{equation}
\normalsize

According to the adversarial training, the perturbation discriminator is trained to distinguish the images reconstructed by the perturbation map generator ($G(x)$) from the real images ($x_r$). Thus, $\mathcal{L}_D$ can be defined as:
\begin{equation}\label{eq:updateD}
\begin{aligned}
    \hspace{-3mm}\mathcal{L}_D = \mathbb{E}_{x_{r} \sim \mathbf{X}_{r}}[\log\left(D\left(x_{r}\right)\right)] +\mathbb{E}_{x \sim \mathbf{X}}[\log\left(1 - D\left(x+G\left(x\right)\right)\right)].
\end{aligned}
\end{equation}
As a result, by alternating the generative adversarial loss and the discriminator loss, the perturbation map generator can generate high-quality perturbation maps.

\subsection{Deepfake Classifier}
The deepfake classifier is a network to distinguish whether the input image is the generated fake one or not.
Thus, the overall framework of the deepfake classifier is a conventional classification network using ResNet-50~\cite{resnet} to predict the binary label for deepfake detection~\cite{frank, adobe}.
We denote the deepfake classifier as $C(\bullet)$.

\subsubsection{Input of Deepfake Classifier}
Since the deepfake classifier detects the presence of the informative features upon the frequency-level artifacts in the input image, the images with the generated perturbation maps should be inserted into the deepfake classifier instead of the raw images.
The input image of the deepfake classifier is defined as:
\begin{equation}
    A_G(x) = x + G(x).
\end{equation}

\begin{algorithm}[t!]
\caption{Training the deepfake detection model}
\label{alg:alg}
\begin{algorithmic}

\State $G,D \gets $ random initial parameters
\State $C \gets $ pre-trained parameters
\State $epoch =0$
\Repeat
      \State $(\mathbf{X}, \mathbf{Y}) \gets $ batch sampled from dataset
      
      \State //Forward Propagation
      \State Estimate $\mathcal{L}_{adv}$, $\mathcal{L}_{com}$, $\mathcal{L}_{D}$, $\mathcal{L}_{C}$ by Eq.~\ref{eq:advLoss},~\ref{eq:comLoss},~\ref{eq:updateD},~\ref{eq:updateC}
      \State //Update parameters according to gradients
      \State Update $G$ by $\arg_G \min_G \mathcal{L}_G$
      \State Update $D$ by $\arg_D \max_D \mathcal{L}_D$
      \State Update $C$ by $\arg_C \min_C \mathcal{L}_C$
      \If{No remaining data}
          \State $epoch \gets epoch + 1$
      \EndIf
\Until $epoch = 20$
\end{algorithmic}
\end{algorithm}

\begin{table*}[t!]
\centering
\scriptsize
\begin{tabular}{lcc|cccccccccccccc}
\hline
\multicolumn{1}{c}{\multirow{2}{*}{Model}} & \multicolumn{2}{c|}{Original} & \multicolumn{2}{c}{Hue} & \multicolumn{2}{c}{Brightness} & \multicolumn{2}{c}{Saturation} & \multicolumn{2}{c}{Gamma} & \multicolumn{2}{c}{Contrast}  & \multicolumn{2}{c}{Blur}& \multicolumn{2}{c}{Rotation}  \\ \cline{2-17} 
 & Acc. & A.P. & Acc. & A.P. & Acc. & A.P. & Acc. & A.P. & Acc. & A.P. & Acc. & A.P. & Acc. & A.P. & Acc. & A.P.  \\ \hline
Wang\shortcite{adobe} & 99.9 & \textbf{100.} & 73.9 & 81.3 & 61.8 & 74.7 & 74.3 & 84.4 & 70.2 & 83.2 & 66.6 & 79.7 & 45.3 & 49.1 & 71.3 & 80.4 \\
Frank~\shortcite{frank} & 99.6 & 99.4 & 85.5 & 97.2 & 84.2 & 97.2 & 91.2 & 98.0 & 85.4 & 97.4 & 84.3 & 96.7 &53.8 & 90.3 & 99.6 & 99.4 \\
Durall~\shortcite{watch_cvpr20} & 99.7 & 99.3 & \textbf{98.6} & 97.9 & 93.6 &88.8 & 98.6 & 97.9 & 97.2 & 95.3 & 94.8 & 91.0 & 50.0 & 53.2 & 98.4 & 97.6 \\
Jeong~\shortcite{wacv} & 99.8 & \textbf{100.} & 85.0 & 92.6 & 89.9 & 90.8 & 96.9 & 99.6 & \textbf{99.7} & \textbf{100.} & 90.8 & 91.6 & 67.1 & 99.2 & 99.0 & 100. \\
Ours & \textbf{100.} & \textbf{100.} & 95.0 & \textbf{99.7} & \textbf{99.5} & \textbf{100}. & \textbf{100.} & \textbf{100.} & 85.5 & 98.6 & \textbf{98.8} & \textbf{100.} & \textbf{98.5} & \textbf{98.5} & \textbf{100.} & \textbf{100.}  \\
\hline
\end{tabular}
\caption{\textbf{Comparison of cross-manipulation performance.} }
\label{tab:face_manipulation}
\end{table*}


\begin{table}[]
\centering
\scriptsize
\resizebox{1.00\linewidth}{!}{%
\begin{tabular}{lcccccccccc}
\hline
\multicolumn{1}{c}{Training}                                           & \multicolumn{10}{c}{Resolutions}                                                                                                                                                                                                                             \\ \hline
\multicolumn{1}{c}{\multirow{2}{*}{Model}} &  \multicolumn{2}{c}{$1024\times1024$}                    & \multicolumn{2}{c}{$512\times512$}                      & \multicolumn{2}{c}{$256\times256$}                      & \multicolumn{2}{c}{$128\times128$}                      & \multicolumn{2}{c}{$64\times64$}                        \\ \cline{2-11} 
\multicolumn{1}{c}{} & \multicolumn{1}{c}{Acc.} & \multicolumn{1}{c}{A.P.} & \multicolumn{1}{c}{Acc.} & \multicolumn{1}{c}{A.P.} & \multicolumn{1}{c}{Acc.} & \multicolumn{1}{c}{A.P.} & \multicolumn{1}{c}{Acc.} & \multicolumn{1}{c}{AP.} & \multicolumn{1}{c}{Acc.} & \multicolumn{1}{c}{A.P.} \\ \hline
Wang~\shortcite{adobe} & 99.9 & \textbf{100.} & 97.6 & 97.3 & 66.1 & 74.4 & 62.6 & 69.4 & 50.4 & 54.9 \\
Frank~\shortcite{frank} & 99.6 & 99.4 & 92.2 & 90.2 & 90.5 & 86.0 & 91.3 & 86.9 & 89.7 & 85.1\\
Durall~\shortcite{watch_cvpr20} & 99.7 & 99.3 & 85.1  & 79.0 & 80.0 & 73.7 & 77.2 & 70.9 & 77.9 & 71.7 \\
Jeong ~\shortcite{wacv}& 99.8 & \textbf{100.} & 97.9 & 99.9 & 97.8 & 99.8 & 89.4 & 96.9 & 59.7 & 62.2
\\
Ours & \textbf{100.} & \textbf{100.} & \textbf{100.} & \textbf{100.} & \textbf{100.} & \textbf{100.} & \textbf{98.0} & \textbf{99.9} & \textbf{95.9} & \textbf{99.4} \\
\hline
\end{tabular}
\vspace{-1.0em}
}
\caption{\textbf{Testing results with variance in resolutions.} }
\label{tab:resize_test}
\end{table}
\begin{figure*}[!t]
\centering
        \includegraphics[width=0.9\linewidth]{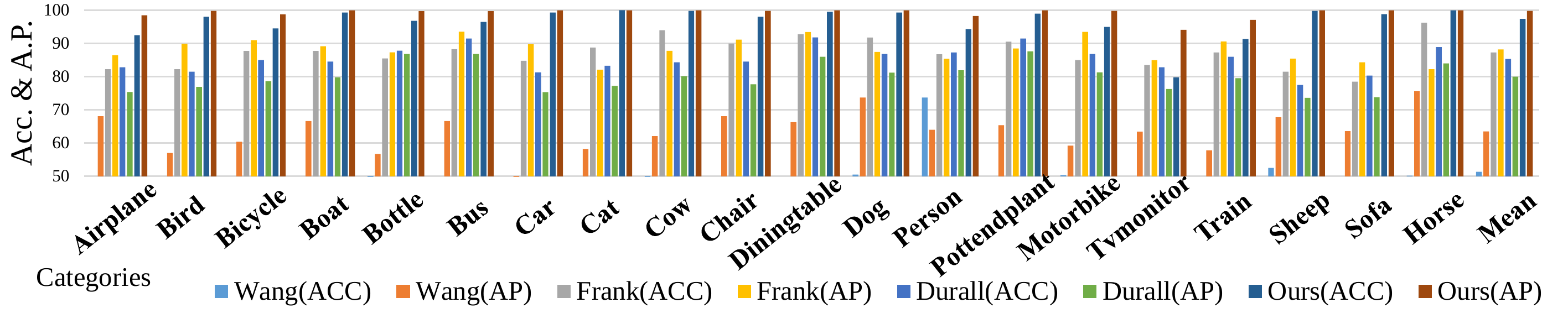}
\caption{\textbf{Comparison of performance in unknown categories.} }
    \label{fig:cross_category}
    \vspace{-1.5em}
\end{figure*}

\subsubsection{Training of Deepfake Classifier}
The training loss of the deepfake classifier is built by the cross-entropy loss as:
\small
\begin{equation}\label{eq:updateC}
\begin{aligned}
    \mathcal{L}_{C}=\mathbb{E}_{(x,y)\sim(\mathbf{X},\mathbf{Y})}& \big[y\log{(C(A_{G}(x)))}\\&+(1-y)\log{(1-C(A_{G}(x)))}\big],
\end{aligned}
\end{equation}
\normalsize
where $\mathbf{Y}$ is the set of real and fake labels paired with the respective samples of $\mathbf{X}$.
Then, the deepfake classifier can be trained as follows: $\widehat{C}=\arg_C \min_C \mathcal{L}_C$.
The training procedure of our overall framework is presented in Algorithm~\ref{alg:alg}.

\subsection{Deepfake Image Prediction}
After the training of our overall framework, we predict whether the new test image is labeled as real or fake by utilizing the perturbation map generator of FrePGAN ($\widehat{G}$) and the deepfake classifier ($\widehat{C}$).
For the new test image $x'$, we acquire the perturbed images by estimating $A_{\widehat{G}} (x')=x'+\widehat{G}(x')$.
Then, $A_{\widehat{G}}(x')$ is fed into the deepfake classifier, which results in the final prediction as: $\widehat{C}(A_{\widehat{G}}(x'))$.

\subsection{Implementation Details}~\label{sec:implementation}
We employ the architecture of VGG model~\cite{vgg} for the perturbation map generator and the discriminator of DCGAN~\cite{dcgan} for the perturbation discriminator.
In addition, we utilize ResNet~\cite{resnet} pre-trained by ImageNet~\cite{imagenet} for the deepfake classifier. 
We use Adam~\cite{adam} to train the perturbation map generator and the perturbation discriminator with the learning rate of $10^{-4}$ and $10^{-1}$, respectively.
Also, the deepfake classifier is trained by Adam~\cite{adam} with the learning rate of $10^{-4}$.
The batch size of the optimizer is always set to $16$, and the input image size is resized to $256\times256$ when the image sizes vary.
The number of epochs is set to $20$.

%% file: 4-result.tex
\begin{table*}[t!]
\centering
\scriptsize
\resizebox{1.00\linewidth}{!}{%
\begin{tabular}{lcccccccccccccccccc|cc}
\hline
\multicolumn{1}{c}{\multirow{3}{*}{Model}} & \multicolumn{2}{c}{Training settings} & \multicolumn{16}{c}{Test Models} \\ \cline{2-21} 
& \multirow{2}{*}{Input} & \multirow{2}{*}{\# class} & \multicolumn{2}{c}{ProGAN} & \multicolumn{2}{c}{StyleGAN} & \multicolumn{2}{c}{StyleGAN2} & \multicolumn{2}{c}{BigGAN} & \multicolumn{2}{c}{CycleGAN} & \multicolumn{2}{c}{StarGAN} & \multicolumn{2}{c}{GauGAN}& \multicolumn{2}{c}{Deepfake} & \multicolumn{2}{|c}{Mean}\\ \cline{4-21} 
 &  &  & Acc. & A.P. & Acc. & A.P. & Acc. & A.P. & Acc. & A.P. & Acc. & A.P.& Acc. & A.P. & Acc. & A.P. & Acc. & A.P. & Acc. & A.P. \\ \hline
Wang~\shortcite{adobe} & Image & 1 & 50.4 & 63.8 & 50.4 & 79.3 & 68.2 & \textbf{94.7} & 50.2 & \textsc{61.3} & 50.0 & 52.9 & 50.0 & 48.2 & 50.3 & 67.6 & 50.1 & 51.5 & 52.5 & 64.9 \\
Frank~\shortcite{frank} & Freq & 1 & 78.9 & 77.9 & 69.4 & 64.8 & 67.4 & 64.0 & 62.3 & 58.6 & 67.4 & 65.4 & 60.5 & 59.5 & 67.5 & 69.1 & 52.4 & 47.3 & 65.7 & 63.3 \\
Durall~\shortcite{watch_cvpr20} & Freq & 1 & 85.1 & 79.5 & 59.2 & 55.2 & 70.4 & 63.8 & 57.0 & 53.9 & 66.7 & 61.4 & \textbf{99.8} & 99.6 & 58.7 & 54.8 & 53.0 & 51.9 & 68.7 & 65.0 \\
Jeong~\shortcite{wacv} & Freq & 1 & 82.5 & 81.4 & 68.0 & 62.8 & 68.8 & 63.6 & \textbf{67.0} & \textbf{62.5} & \textbf{75.5} & \textbf{74.2} & 90.1 & 90.1 & \textbf{73.6} & \textbf{92.1} & 51.6 & 49.9 & 72.1 & 72.1 \\
Our & Image & 1 & \textbf{95.5} & \textbf{99.4} & \textbf{80.6} & \textbf{90.6} & \textbf{77.4} & 93.0 & 63.5 & 60.5  & 59.4 & 59.9 
& 99.6 & \textbf{100.} & 53.0 & 49.1 & \textbf{70.4} & \textbf{81.5} & \textbf{74.9} & \textbf{79.3}  \\
\hline
Wang~\shortcite{adobe} & Image  & 2 & 64.6 & 92.7 & 52.8 & 82.8 & 75.7 & \textbf{96.6} & 51.6 & 70.5 & 58.6 & 81.5 & 51.2 & 74.3 & 53.6 & 86.6 & 50.6 & 51.5 & 57.3 & 79.6 \\
Frank~\shortcite{frank} & Freq & 2 & 85.7 & 81.3 & 73.1 & 68.5 & 75.0 & 70.9 & 76.9 & 70.8 & \textbf{86.5} & 80.8 & 85.0 & 77.0 & 67.3 & 65.3 & 50.1 & 55.3 & 75.0 & 71.2  \\
Durall~\shortcite{watch_cvpr20} & Freq & 2 & 79.0 & 73.9 & 63.6 & 58.8 & 67.3 & 62.1 & 69.5 & 62.9  & 65.4 & 60.8 & \textbf{99.4} & 99.4 & 67.0 & 63.0 & 50.5 & 50.2 & 70.2 & 66.4 \\
Jeong~\shortcite{wacv} & Freq & 2 & 87.4 & 87.4 & 71.6 & 74.1 & \textbf{77.0} & 81.1 & \textbf{82.6} & \textbf{80.6} & 86.0 & \textbf{86.6} & 93.8 & 80.8 & \textbf{75.3} & \textbf{88.2} & 53.7 & 54.0 & \textbf{78.4} & 79.1 \\

Our & Image & 2 & \textbf{99.0} & \textbf{99.9} & \textbf{80.8} & \textbf{92.0} & 72.2 & 94.0 & 66.0 & 61.8 & 69.1 & 70.3 & 98.5 & \textbf{100.} & 53.1 & 51.0 & \textbf{62.2} & \textbf{80.6} & 75.1 & \textbf{81.2}\\
 \hline
Wang\shortcite{adobe} & Image  & 4 & 91.4 & 99.4 & 63.8 & \textbf{91.4} & 76.4 & 97.5 & 52.9 & 73.3 & 72.7 & \textbf{88.6} & 63.8 & 90.8 & 63.9 & \textbf{92.2} & 51.7 & 62.3 & 67.1 & 86.9 \\
Frank~\shortcite{frank} & Freq & 4 & 90.3 & 85.2 & 74.5 & 72.0 & 73.1 & 71.4 & \textbf{88.7} & \textbf{86.0} & 75.5 & 71.2 & 99.5 & 99.5 & 69.2 & 77.4 & 60.7 & 49.1 & 78.9 & 76.5 \\
Durall~\shortcite{watch_cvpr20} & Freq & 4 & 81.1 & 74.4 & 54.4 & 52.6 & 66.8 & 62.0 & 60.1& 56.3  &  69.0 & 64.0 & 98.1 & 98.1 & 61.9 & 57.4 & 50.2 & 50.0 & 67.7 & 64.4 \\
Jeong~\shortcite{wacv} & Freq & 4 & 90.7 & 86.2 & 76.9 & 75.1 & 76.2 & 74.7 & 84.9 & 81.7 & \textbf{81.9} & 78.9 & 94.4 & 94.4 & \textbf{69.5} & 78.1 & 54.4 & 54.6 & 78.6 & 77.9 \\
Our & Image & 4 & \textbf{99.0} & \textbf{99.9} & \textbf{80.7} & 89.6 & \textbf{84.1} & \textbf{98.6} & 69.2 & 71.1 & 71.1 & 74.4 & \textbf{99.9} & \textbf{100.} & 60.3& 71.7 & \textbf{70.9} & \textbf{91.9} & \textbf{79.4} & \textbf{87.2}  \\

\hline
\vspace{-1.5em}
\end{tabular}
}
\caption{\textbf{Comparison of cross-model performance.} }
\label{tab:cross_gan}
\end{table*}

\begin{table*}[]
\centering
\scriptsize
\begin{tabular}{ccc|cccccccc|cc}
\hline
\multicolumn{3}{c|}{\multirow{1}{*}{Ablation settings}} & \multicolumn{2}{c}{Self} & \multicolumn{2}{c}{Category} & \multicolumn{2}{c}{Model} & \multicolumn{2}{c|}{Manipulation}  & \multicolumn{2}{c}{Mean} \\
\cline{1-13} 
Generator & $L_{com}$ & $L_{adv}$ & Acc.  & A.P.      & mAcc.   & mA.P.  & mAcc.     & mA.P.     & mAcc.  & mA.P. & Acc.  & A.P.   \\
\hline
 Freq &  & \checkmark & 99.0 & 100. & 90.9 & 98.2 & 69.4 & 76.1 & 90.9 & 98.4 & 87.6 & 93.2  \\
Freq & \checkmark &  & 100. & 100. & 91.0 & 98.5 & 67.3 & 74.4 & 91.3 & 99.1 & 87.4 & 93.0 \\
 Img & \checkmark & \checkmark & 99.8 & 100. & 92.9 & 99.2 & 68.7 & 77.4 & 92.4 & 99.6 & 88.5 & 94.1 \\

Freq & \checkmark & \checkmark & 100. & 100. & 95.5 & 99.4 & 74.9 & 79.3 & 96.8 & 99.5 & 91.8 & 94.6 \\

\hline
\end{tabular}
\caption{\textbf{Ablation Test with Various Settings.}}
\label{tab:ablation_study}
\vspace{-1.5em}
\end{table*}

\section{Experimental Results}
We conduct experiments to confirm the performance of the deepfake detector in the known domain and unseen domain.

\subsection{Dataset}
We conduct experiments based on the same trainset and testset of the experimental data of Wang~\etal~\cite{adobe}. The trainset contains 20 objects of Progan~\cite{progan}. The testset consists of  FFHQ~\cite{stylegan} and LSUN~\cite{lsun} to train ProGAN~\cite{progan}, StyleGAN~\cite{stylegan}, and StyleGAN2~\cite{stylegan2}, 
and employs Imagenet~\cite{imagenet} to train BigGAN~\cite{biggan} and CycleGAN~\cite{cyclegan}. 
Also, we use CelebA~\cite{celeba} for training StarGAN~\cite{stargan}, and COCO~\cite{coco} for training GauGAN~\cite{gaugan}. 
Lastly, we utilize Deepfake dataset~\cite{faceforensics++}, which is a combination of various videos collected online with partially generated images reconstructed by face-swapping models.

Also, to test the model's performance in various manipulation techniques and resizing, we employ the face data of ProGAN~\cite{progan} dataset in $1,024\times1,024$ resolution. For the experiments with unknown categories and unknown models, we utilize the horse data of ProGAN~\cite{progan} dataset in $256\times256$ resolution.

\subsection{Deepfake Detection Performance}
The deepfake detection performance is tested by the four types of experiments: manipulated face images, resized face images, unseen categories, and unseen models.
We utilize two evaluation metrics of the average precision score (A.P.) and accuracy (Acc.) as represented by \cite{adobe, watch_cvpr20, frank}.
To validate the effectiveness of the proposed deepfake detector, we select the image-based~\cite{adobe} and frequency-based state-of-the-art models~\cite{frank, watch_cvpr20,wacv} for comparison.

\subsubsection{Deepfake Detection of Manipulated Face Images}
We conduct various image manipulation experiments using the face data of ProGAN~\cite{progan} in $1,024\times1,024$ resolution. 
To test the performance with unknown manipulations, we add 7 various changes in images, such as adjusting the hue, brightness, saturation, gamma, contrast, blurriness, and image rotation. 
As shown in Table~\ref{tab:face_manipulation}, ours is the most robust model achieving superior performance in image manipulation experiments. 

\subsubsection{Deepfake Detection of Resized Face Images}
To test the model with the previous detectors' chronic issue of significant performance decline with resizing, we conduct experiments with the face data of ProGAN~\cite{progan} dataset by gradually reducing the image sizes with five different resolutions from $1,024\times1,024$ to $64\times64$. 
Based on the experimental results of the resizing performance of the models as shown in Table~\ref{tab:resize_test}, we can confirm that our model outperforms all other models when tested with the five cases of resized resolutions. 
Furthermore, even when the image resolution is reduced, our model maintains 100\% performance from $1,024\times1,024$ to $256\times256$. When reduced to $128\times128$ and $64\times64$, our model's performance slightly declines but maintains at least 97.8\%, proving the best performance compared to the existing model.

\subsubsection{Deepfake Detection of Unknown Categories}
As shown in Figure~\ref{fig:cross_category}, we conduct various experiments using the three classifiers to analyze the performance of the models with 20 unknown categories. 
We compare our model's performance to the previous state-of-the-art models~\cite{adobe,watch_cvpr20,frank}. 
The experimental results verify that ours is the most robust model in all categories, even when the number of training classes increases and the type of inputs varies.
The variety in the testing environment shows that each component in our model plays an important role to detect not only the test category but also all categories.

\subsubsection{Deepfake Detection of Unknown GAN Models}
To expand the testing scope, we compare the performance of our model with 8 different generative models, including ProGAN~\cite{progan}, StyleGAN~\cite{stylegan}, StyleGAN2~\cite{stylegan2}, BigGAN~\cite{biggan}, CycleGAN~\cite{cyclegan}, StarGAN~\cite{stargan}, GauGAN~\cite{gaugan}, and Deepfake~\cite{faceforensics++}. 
As shown in Table~\ref{tab:cross_gan}, we make changes to the training settings and conduct experiments to detect the GAN models. First, we train the models with one type of category and test with all GAN models. Then, to add variety, we increase the number of training categories to two and four. 
Interestingly, even when trained with only one type of category, our model achieves excellent performance similar to the case when trained with four classes.
The results show that ours achieves the highest performance in both Acc. and A.P in ProGAN~\cite{progan}, StyleGAN~\cite{stylegan}, GauGAN~\cite{gaugan}, and Deepfake~\cite{faceforensics++}. 
Also, in StyleGAN2~\cite{stylegan2}, BigGAN~\cite{biggan}, CycleGAN~\cite{cyclegan}, and StarGAN~\cite{stargan}, our model achieves the best performance in either Acc. or A.P.  
Our performance rises with the number of training categories, even when tested with the partially generated images in Deepfake dataset.

\begin{figure}[t]
\centering
\includegraphics[width=\linewidth]{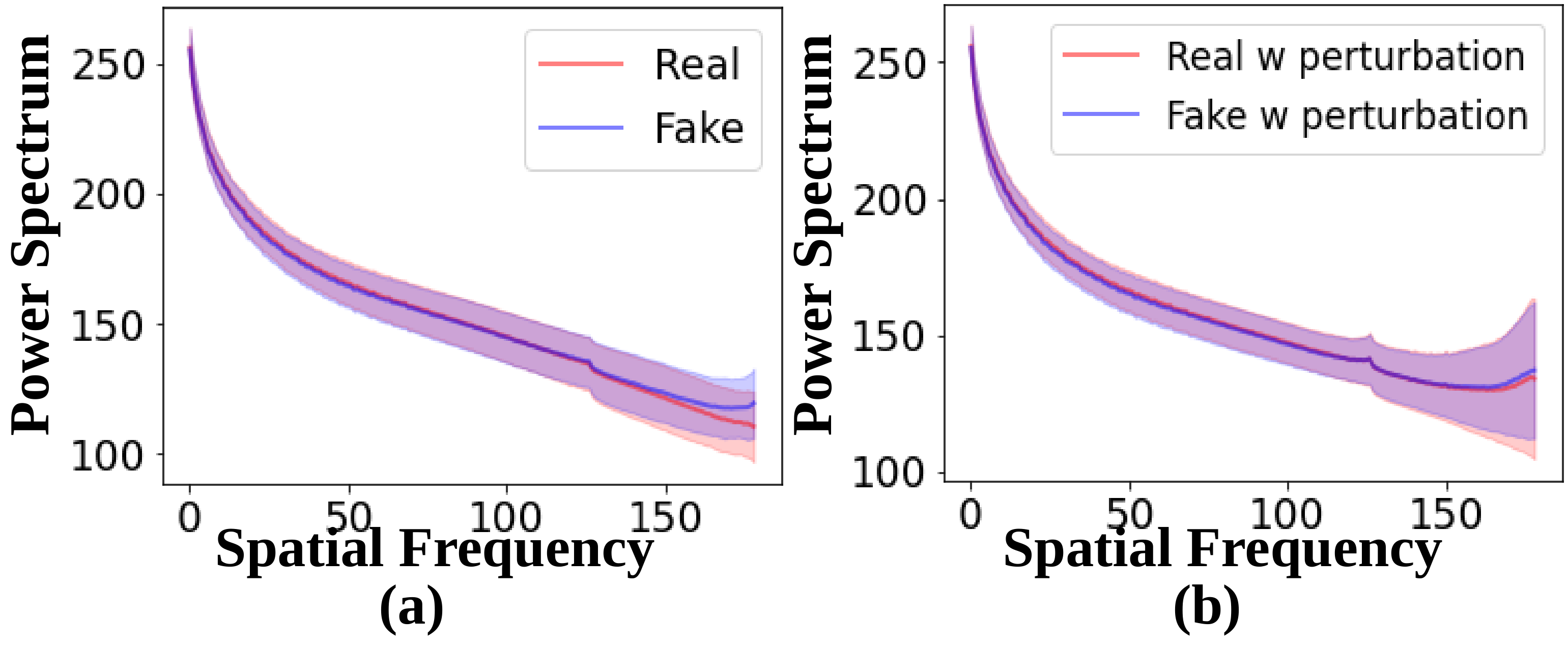}
\caption{\textbf{Addition of Perturbation Maps.} (a) shows the averaged power spectrum of real and fake images, respectively. (b) compares the averaged power spectrum of real with that of fake. While the real and fake images can be distinguished by the high-frequency components in the power spectrum, the difference is nearly removed after the addition of the generated perturbation maps.}
\label{fig:perb_spec} 
\vspace{-1.5em}
\end{figure}

\begin{figure}[t]
\centering\vspace{-1.0em}
        \includegraphics[width=0.97\linewidth]{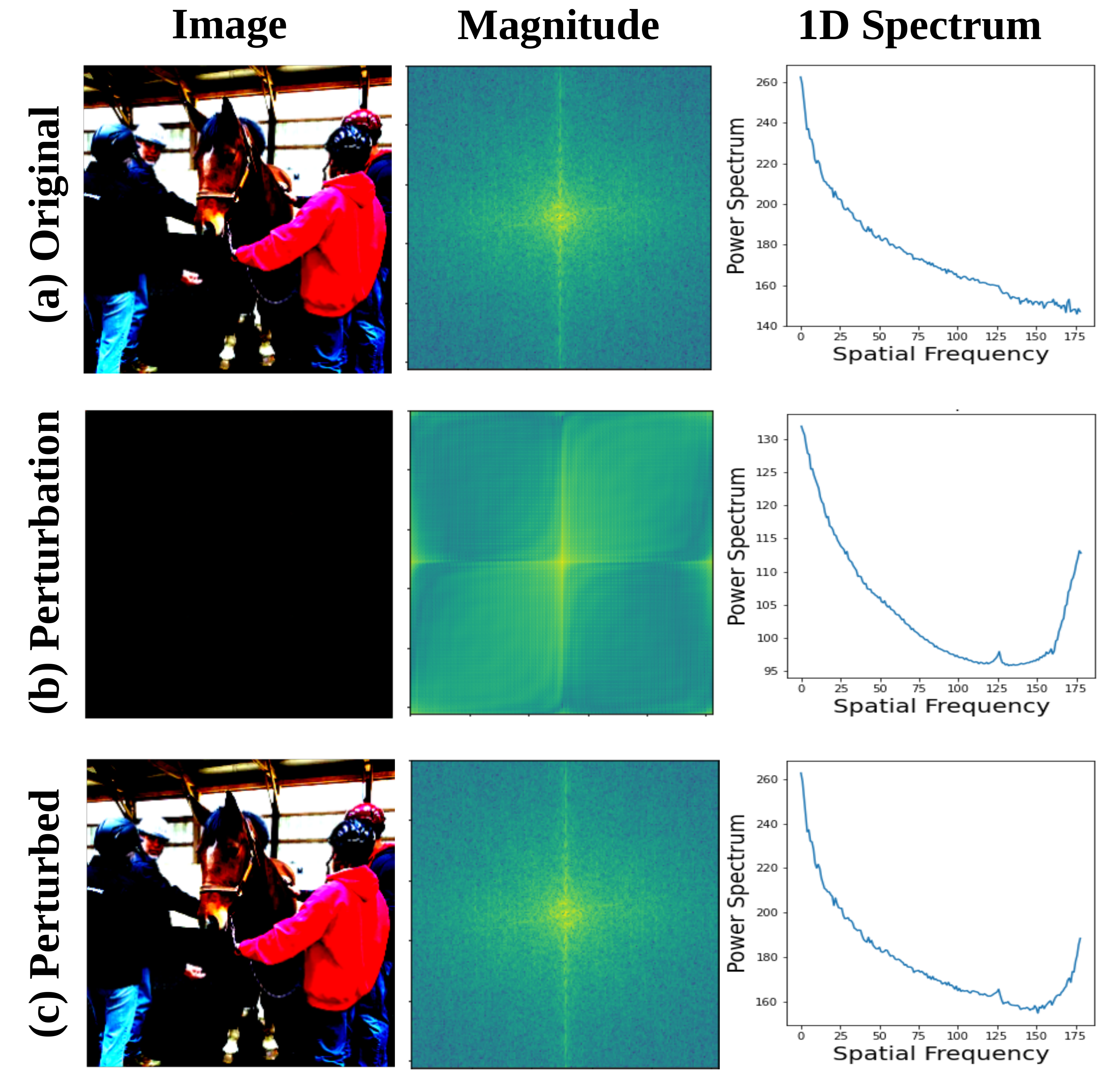}
\caption{\textbf{Effect of Perturbation Map.} With the compression loss, the magnitude of perturbation maps is negligent in the pixel-level domain as shown in (b). However, the perturbation map increments the high-frequency components as shown in (c).}
    \label{fig:add_power_spectrum}
    \vspace{-1.5em}
\end{figure}

\subsection{Ablation Study}
To validate the effectiveness of the components in the proposed framework, we also test several variants.
We modify four settings of the proposed framework, which include the composition of the perturbation generation loss and the data input types of the generator of FrePGAN.
The training dataset for the ablation tests is the horse images generated by ProGAN~\cite{progan}. The testing environments for cross-category and cross-model experiments are the same as the previous section, and the manipulation experiment is conducted by manipulating the horse test images with the same methods of cross-manipulation experiments.

The results of the ablation tests are provided in Table~\ref{tab:ablation_study}.
The best performance of $(91.8,~94.6)$ can be obtained by utilizing the entire framework with the frequency-level generator, the adversarial learning mechanism, and the compression loss.
Interestingly, the simultaneous usage of the compression loss and the adversarial loss makes synergy to improve the generalization of deepfake classifier.
This result shows that the quality of perturbation is important for the overall performance.
Also, when we replace our frequency-level generator with the pixel-level generator without the frequency transformers, the performance drops by far due to the limited quality of the generated perturbation maps.

\subsection{Visualization of Perturbation Maps}
To visualize the effect of the perturbation maps, we conduct two experiments. First, as shown in Fig.~\ref{fig:perb_spec}, we obtain the power spectra of real and fake images. Then, due to the frequency-level artifacts, the high-frequency components of real and fake become distinct to distinguish between them. However, after adding the perturbation maps, the difference is reduced by far in the power spectra, which validates that the proposed framework successfully generates high-quality perturbation maps to make the real and fake similar.

Second, as shown in Fig.~\ref{fig:add_power_spectrum}, from one real image, we obtain both the 1D and 2D power spectrum. Interestingly, even after the addition of perturbation maps, the pixel-level image and its 2D power spectrum are almost preserved. However, when we estimate the 1D power spectrum, we can find that the high-frequency components are magnified, which results in the 1D power spectrum similar to that of fake images containing the frequency-level artifacts.

%% file: 5-conclusion.tex
\section{Conclusion}
It has become highly important to develop a robust, generalized deepfake detector, which is not limited to the training settings. 
Numerous experiments validate that our framework achieves a generalized detection robust in various testing scenarios including the unknown categories, GAN models, manipulations, and resizing. 
Trained with the perturbation generation loss and compression loss, our newly proposed FrePGAN generates perturbations to reduce the effects of domain-specific artifacts in generated images. 
Also, our framework shows the effectiveness of the alternate updates of the deepfake classifier and the perturbation generator, which is validated to be helpful for the improved generalization of deepfake detectors.